\documentclass[conference]{IEEEtran}
\IEEEoverridecommandlockouts

\usepackage{amsmath,amssymb,amsfonts}
\usepackage{algorithmic}
\usepackage{graphicx}
\usepackage{textcomp}
\usepackage{xcolor}

\definecolor{lamarckbg}{RGB}{220,240,255}
\definecolor{noveltybg}{RGB}{225,255,225}
\definecolor{darkgreen}{RGB}{0,100,0}

\usepackage[most]{tcolorbox}
\newtcolorbox{algbox}[2][]{
  colback=#2!10,
  colframe=#2!60!black,  
  boxrule=0.75pt,
  arc=2pt,
  left=4pt,
  right=4pt,
  top=2pt,
  bottom=2pt,
  width=0.8\linewidth,
  #1
}
\newtcbox{\legendtcbox}[1][]{  
  on line,
  boxrule=0.75pt,
  arc=1pt,
  colback=gray!10,
  colframe=gray!60,
  left=2pt,
  right=2pt,
  top=1pt,
  bottom=1pt,
  #1
}

\usepackage{booktabs}
\usepackage{multirow}
\def\BibTeX{{\rm B\kern-.05em{\sc i\kern-.025em b}\kern-.08em
    T\kern-.1667em\lower.7ex\hbox{E}\kern-.125emX}}

\usepackage{subcaption}
\usepackage[ruled,vlined,linesnumbered]{algorithm2e}
\SetKwInput{KwParams}{Parameters} 
\usepackage[backend=biber, style=ieee, sorting=none]{biblatex}
\usepackage{hyperref}
\begin{document}



\title{Limits of Lamarckian Evolution Under Pressure of Morphological Novelty

} 

\author{\IEEEauthorblockN{Jed R Muff}
\IEEEauthorblockA{\textit{Department of Computer Science} \\
\textit{Vrije Universiteit}\\
Amsterdam, The Netherlands \\
0009-0004-9934-6403}
\and
\IEEEauthorblockN{Karine Miras}
\IEEEauthorblockA{\textit{Department of Computer Science} \\
\textit{Vrije Universiteit}\\
Amsterdam, The Netherlands \\
0000-0003-4942-3488}
\and
\IEEEauthorblockN{A.E. Eiben}
\IEEEauthorblockA{\textit{Department of Computer Science} \\
\textit{Vrije Universiteit}\\
Amsterdam, The Netherlands \\
0000-0002-3106-4213}
}

\maketitle

\begin{abstract}
Lamarckian inheritance has been shown to be a powerful accelerator in systems where the joint evolution of robot morphologies and controllers is enhanced with individual learning. Its defining advantage lies in the offspring inheriting controllers learned by their parents.
The efficacy of this option, however, relies on morphological similarity between parent and offspring. In this study, we examine how Lamarckian inheritance performs when the search process is driven toward high morphological variance, potentially straining the requirement for parent-offspring similarity. Using a system of modular robots that can evolve and learn to solve a locomotion task, we compare Darwinian and Lamarckian evolution to determine how they respond to shifting from pure task-based selection to a multi-objective pressure that also rewards morphological novelty. Our results confirm that Lamarckian evolution outperforms Darwinian evolution when optimizing task-performance alone. However, introducing selection pressure for morphological diversity causes a substantial performance drop, which is much greater in the Lamarckian system. Further analyses show that promoting diversity reduces parent-offspring similarity, which in turn reduces the benefits of inheriting controllers learned by parents. These results reveal the limits of Lamarckian evolution by exposing a fundamental trade-off between inheritance-based exploitation and diversity-driven exploration.

\end{abstract}

\begin{IEEEkeywords}
Evolutionary Robotics, Lamarckian Evolution, Limits of Lamarckian Evolution, Morphological Evolution, Evolution and Learning
\end{IEEEkeywords}

\section{Introduction} \label{sec:intro}

Evolutionary robotics aims to automatically design robot morphologies and control systems through artificial evolution, mimicking natural selection to discover effective solutions without extensive human engineering \cite{eiben2015grand}. A central challenge in this field is the joint optimization of body plans and controllers, a problem compounded by the vast search space and the need for each candidate robot to undergo costly learning or adaptation during its `lifetime' before its evolutionary fitness can be assessed.

Lamarckian evolution offers a compelling mechanism to accelerate this process by allowing offspring to inherit learned traits from their parents. In evolutionary robotics, this typically takes the form of \emph{warm starting}: initializing an offspring’s controller with parameters learned by its parent during lifetime learning, rather than starting from scratch \cite{jelisavcic2019lamarckian}. A substantial body of work has shown that such Lamarckian mechanisms can significantly speed up learning and improve evolutionary performance across a range of robotic systems and learning algorithms \cite{jelisavcic2019lamarckian,jelisavcic2017analysis,jelisavcic2018morphological,jelisavcic2017benefits,luo2023enhancing,luo2023comparative,luo2023comparison,luo2025lamarckian}.

However, the effectiveness of Lamarckian inheritance relies on a critical assumption: that offspring are sufficiently similar to their parents for inherited controllers to remain compatible. Jelisavcic et al.~\cite{jelisavcic2019lamarckian} demonstrated that the benefits of controller inheritance strongly depend on parent--offspring morphological similarity, with performance gains diminishing as morphological divergence increases. In many prior studies, this assumption is implicitly satisfied, as evolution tends to converge toward relatively homogeneous morphologies over time.

Another important factor shaping the limits of Lamarckian evolution is environmental variability. Learning typically adapts agents to local, transient conditions, whereas evolution operates on longer time scales and targets more persistent environmental structure. As a result, inheriting a controller adapted to a previous environment can be counterproductive when conditions change. This intuition is supported by recent work showing that following certain environmental transitions, particularly shifts from easier to more challenging tasks, Lamarckian populations can temporarily underperform Darwinian ones \cite{luo2025lamarckian}. These findings suggest that Lamarckian inheritance is most effective in settings where disruptive forces, such as abrupt environmental change, are limited.

Taken together, this evidence indicates that Lamarckian evolution performs best in regimes characterized by relative stability, both in morphology and environment. We interpret this as suggesting that Lamarckism’s strength relies on the absence of forces that introduce substantial transgenerational variability. Such forces may include environmental non-stationarity, as well as evolutionary pressures that explicitly violate the parent--offspring similarity assumption under which controller inheritance is effective.

In this work, we focus on the latter. Specifically, we hypothesize that \emph{explicitly selecting for morphological diversity will reduce the benefits of Lamarckian evolution, while leaving the benefits of Darwinian evolution largely unaffected}. Although parent--offspring similarity is advantageous in any evolutionary system, it is particularly critical in the Lamarckian case, where inherited controllers must remain compatible with the offspring’s body.

To test this hypothesis, we present a systematic study comparing Darwinian and Lamarckian evolution under differing evolutionary pressures. We evaluate four conditions: Darwinian versus Lamarckian inheritance, each combined with either a simple locomotion objective or an objective that additionally rewards morphological diversity. By analyzing population fitness and learning dynamics, our goal is to determine whether the benefits of Lamarckian inheritance critically depend on morphological similarity, or whether they persist under evolutionary pressures that actively promote morphological change.

The code for this work is available at: \href{https://tinyurl.com/mpfak3xt}{https://tinyurl.com/mpfak3xt}. 
Videos showcasing interesting evolved individuals under various experimental conditions can be viewed here:
\href{https://tinyurl.com/3rj3w27k}{https://tinyurl.com/3rj3w27k}.

\section{Related Work}
\label{sec:related}

Lamarckian inheritance has been extensively studied in evolutionary robotics as a mechanism for accelerating learning through controller transfer. Early and contemporary work has consistently shown that initializing offspring controllers with parameters learned by their parents can lead to faster convergence and improved performance in static environments \cite{mingo2013lamarckism,sasaki2000comparison,jelisavcic2017benefits}. More recent studies have extended these findings to modular robots and complex co-evolutionary settings, further demonstrating the potential of Lamarckian evolution to reduce lifetime learning costs \cite{jelisavcic2018morphological,jelisavcic2019lamarckian}.

A few studies have also examined Lamarckian mechanisms under non-stationary conditions. Luo et al.~\cite{luo2025lamarckian} investigated environmental change and showed that although Lamarckian populations may initially suffer after abrupt task transitions, they often recover more quickly and achieve higher long-term performance than Darwinian populations. Similar trade-offs between rapid exploitation and adaptability have been observed in earlier evolutionary computation work, where Lamarckian inheritance was found to reduce robustness in dynamic environments \cite{sasaki2000comparison}.

Beyond evolutionary robotics, Lamarckian principles have been revisited in modern learning systems. Applications include neural architecture search \cite{sharifi2024lcodeepneat} and soft robotic co-design \cite{tack2024lamarckian}, where variants of weight inheritance and transfer learning are used to mitigate mismatches between inherited controllers and changing morphologies or architectures. These approaches demonstrate that Lamarckian ideas remain relevant, but often rely on mechanisms specifically designed to preserve functional continuity across structural change.

In parallel, a substantial body of work in evolutionary robotics has focused on promoting morphological diversity. Novelty search and quality-diversity algorithms explicitly encourage exploration of diverse body plans, often leading to more creative, adaptable, or evolvable designs \cite{lehman2011abandoning,pugh2016quality,cully2015robots}. However, morphological variation can disrupt learned behaviors, and prior studies have shown that evolvability depends on both environmental structure \cite{auerbach2014environmental} and intrinsic morphological properties \cite{cheney2013unshackling}. These approaches typically assume Darwinian inheritance and do not incorporate lifetime learning or controller transfer.

Despite extensive work on Lamarckian inheritance and on morphological diversity in isolation, their interaction has received little direct attention. Existing studies suggest that Lamarckian benefits rely on parent--offspring similarity, but it remains unclear whether these benefits persist when evolution actively promotes morphological diversity. Our work addresses this gap by explicitly combining diversity pressure with Lamarckian inheritance, enabling a systematic evaluation of warm starting under conditions where parent--offspring similarity is intentionally weakened.

\section{Methods}
\subsection{Robot Platform and Task Definition}

\begin{figure}
    \centering
    \includegraphics[width=\linewidth]{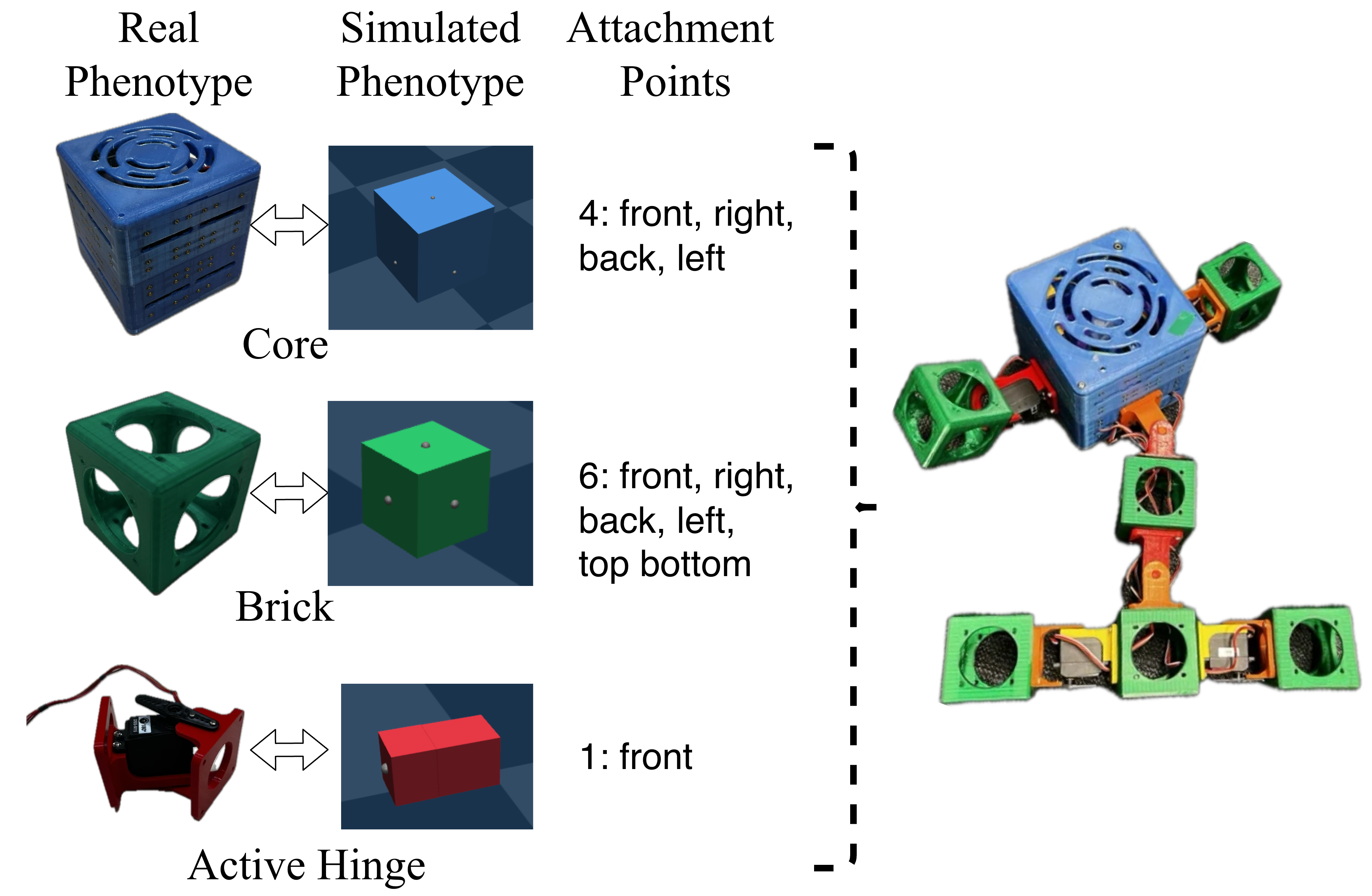}
    \caption{ARIEL framework phenotype space including three components, two passive and one active which is controlled by the brain. On the right demonstrates what a single individual would look like in real life.}
    \label{fig:phenotype}
\end{figure}
We use the ARIEL framework (previously known as Revolve) similar to that used in previous works \cite{jelisavcic2019lamarckian,luo2025lamarckian,luo2023comparison} for simulated modular robots, implemented in the MuJoCo physics simulator. Each robot consists of a fixed core module and a variable number of rigid brick modules and active hinge joints. Figure \ref{fig:phenotype} shows what each component looks like in real life and in simulation and how these parts can be combined together to make an individual. 

The locomotion task requires robots to maximize forward displacement on a flat terrain over a 30-second period. The evaluation proceeds in two phases: first, a 5-second settling phase where the robot is spawned above ground and the brain sends no commands, allowing the robot to settle; then, a 30-second active phase where the brain controls the robot's actuators. The fitness function is calculated as the distance traveled by the robot's core module from its starting position along the positive x-axis, with two penalty terms applied. First, a penalty is applied based on the robot's initial height (measured before the settling phase begins) to discourage designs that exploit high starting positions. Second, a penalty of 0.005 is applied for each instance in which the hinge part touches the ground during evaluation, steering the evolutionary process away from behaviors that exploit physics glitches in simulation. If the hinge part touches the ground more than 200 times during a single evaluation, the fitness is set to -1 regardless of distance traveled, as this indicates physics exploitation for forward propulsion.

\begin{equation}
    f = \begin{cases}
-1 & \text{if } c_{\text{hinge}} > 200 \\
d - l - 0.005 \cdot c_{\text{hinge}} & \text{otherwise}
\end{cases}
\end{equation}

\noindent where $f$ is the fitness value, $d$ is the Euclidean distance traveled by the robot's core module, $l$ is the penalty based on the initial height, $c_{\text{hinge}}$ is the number of contacts between the hinge parts and the ground.
\subsection{Evolutionary Algorithm}
We use the ($mu+lambda$) evolution strategy as outlined in Algorithm \ref{alg:mu+lambda}. Some important details to highlight include the following. 80\% of the population uses the mutation operator, while 20\% uses the crossover operator based on initial parameter tuning. The initial population $P_0$ consists of 10 random generated small trees (depth 2), 10 medium trees (depth 3), and 10 large trees (depth 4) randomly generated. After crossover, the offspring undergo the mutation operator. The weight inheritance of the artificial neural network during crossover comes from the closest parent -- determined using the tree edit distance regarding the body. This ensures that the offspring acquire the brain that is most likely to work for its new body. Although fitness is evaluated once per individual, novelty scores are recomputed at each generation for all individuals in the population and offspring pool.

\SetKwInput{KwInput}{Input}
\SetKwInput{KwParams}{Parameters}
\begin{algorithm}
\caption{($\mu+\lambda$) Evolution Strategy. Optional components for \protect\legendtcbox[colback=lamarckbg,colframe=lamarckbg!30!blue]{Lamarckian inheritance} and \legendtcbox[colback=noveltybg,colframe=noveltybg!30!darkgreen]{novelty search} are shown as shaded boxes; these blocks are included only in the corresponding algorithm variants.} \label{alg:mu+lambda}
\KwParams{$\mu{=}30$, $\lambda{=}30$, $n_{\text{mut}}{=}24$, $n_{\text{cross}}{=}6$}

$P_0 \gets$ Random population of $\mu$ tree genotypes 

EvaluatePopulation($P_0$)

\begin{algbox}[width=.4\linewidth]{green}
ComputeNovelty($P_0$)
\end{algbox}

\For{$g = 1 \to n_{\text{gen}}$}{
    $O_{\text{mut}} \gets$ Mutate(TournamentSelect($P_{g-1}$, $n_{\text{mut}}$))
    
    $O_{\text{cross}} \gets$ Crossover(TournamentSelect($P_{g-1}$, $2n_{\text{cross}}$))
    
    $O_g \gets O_{\text{mut}} \cup O_{\text{cross}}$

    \begin{algbox}[width=.7\linewidth]{blue}
    \For{$c \in O_g$}{
        InheritWeights($c$, ClosestParent($c$))
    }
    \end{algbox}
    EvaluatePopulation($O_g$)
    \begin{algbox}[width=.7\linewidth]{green}
    UpdateNoveltyArchive($P_{g-1}{~}{\cup}{~}O_g$)
    UpdateFitnessWithNovelty($P_{g-1} \cup O_g$)
    \end{algbox}
    
    $P_g \gets \text{Top}_\mu(P_{g-1} \cup O_g)$
}

\end{algorithm}
\subsection{Representation and Reproduction}
\begin{figure}
    \centering
    \includegraphics[width=0.95\linewidth]{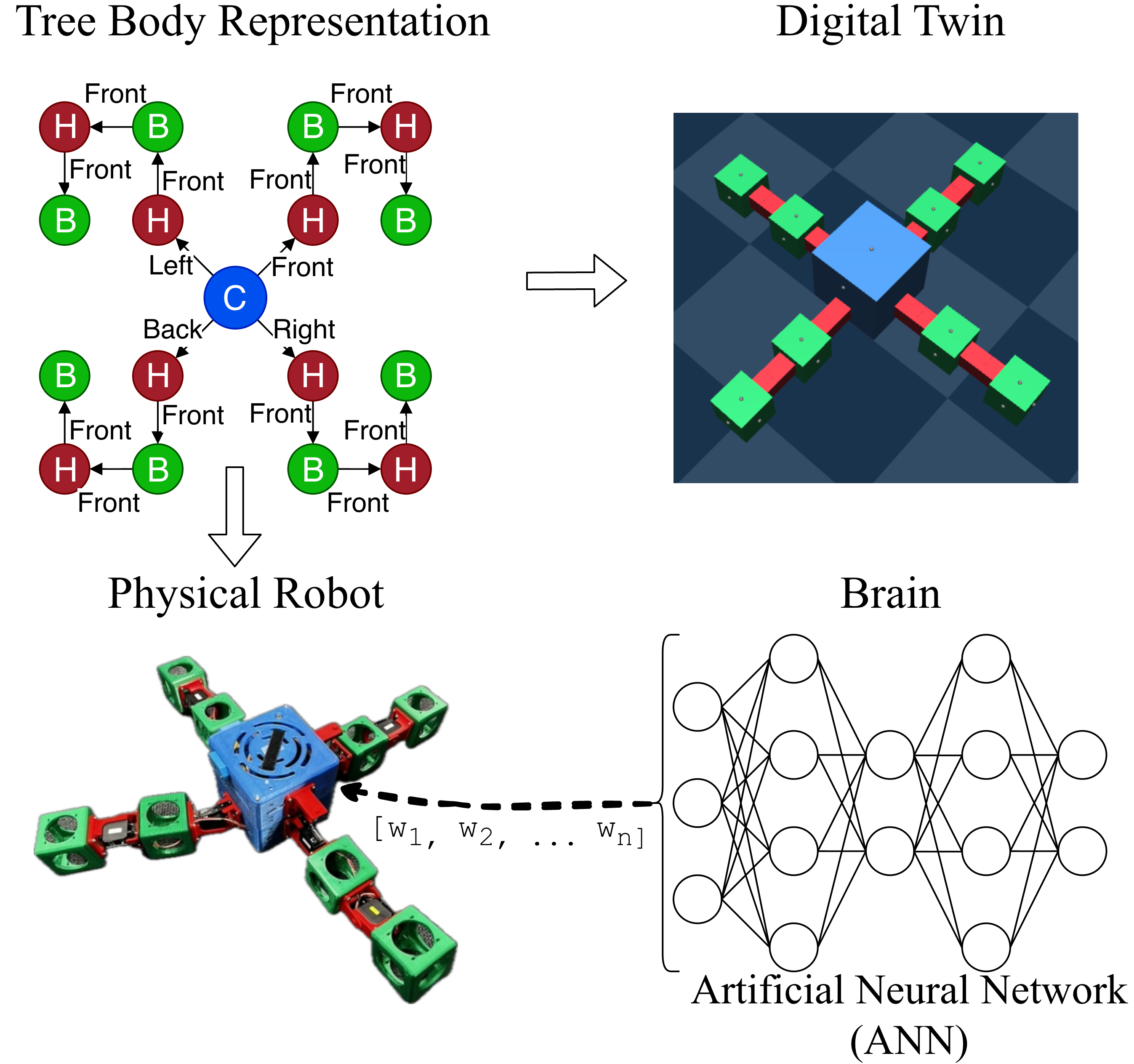}
    \caption{Representation used in this work. The body employs a tree-based representation, while the brain uses an ordered array of parameters that maps to an artificial neural network (ANN). Note that the ANN depicted in this diagram is illustrative only and does not reflect the actual ANN architecture implemented in this paper (see Section \ref{sec:brain}).}
    \label{fig:representation}
\end{figure}
\subsubsection{Body}
The body is encoded as a tree structure rooted at the core, with nodes corresponding to modules and edges representing physical connections (Figure \ref{fig:phenotype}). This tree simultaneously defines the body genotype and is evolved using standard tree-based genetic operators, namely mutation and crossover. To ensure morphological feasibility and computational tractability, several constraints are imposed. First, the total number of body parts is capped at 25, with at most 12 active hinges. If a genetic operation produces a morphology exceeding this hinge limit, excess hinges are pruned using a breadth-first traversal of the tree. These constraints limit morphological complexity while maintaining reasonable simulation times. Second, a minimum of four hinges is required; morphologies that do not satisfy this condition are assigned a fitness of zero, preventing computational resources from being spent on infeasible or degenerate solutions.

To ensure physical realizability, all generated tree genotypes, whether produced through random initialization, mutation, or crossover, undergo a pruning procedure to eliminate colliding components. This pruning is implemented via breadth-first search traversal of the genotype tree. During traversal, the morphology is incrementally constructed using the Python-FCL collision detection library, with each prospective body part being evaluated for spatial interference with all previously placed components. When a collision is detected, the offending node and all of its descendant nodes are removed from the tree, ensuring that only physically valid morphologies are retained for evaluation.

\subsubsection{Brain} \label{sec:brain}

The brain phenotype is an artificial neural network (ANN) with three hidden layers containing 32, 16, and 32 neurons, respectively. This architecture was determined through hyperparameter optimization conducted on a reference gecko morphology with 6 hinges. The input layer consists of 9 + h neurons, where h denotes the number of hinges in the morphology. These inputs encode the core's Euler orientation (3 values), linear velocity (3 values), angular velocity (3 values), and the current joint positions for all h hinges. The output layer contains h neurons, each specifying the target position of its corresponding joint. 

This phenotype is represented in genotype space as one long ordered array of parameters. The total number of trainable parameters (weights and biases) in the network is given by:

\begin{equation}
    \begin{aligned}
        P = (n_0 \times n_1 + n_1) + (n_1 \times n_2 + n_2) \\+ (n_3 \times n_3 + n_3) + (n_3 \times n_4 + n_4)
    \end{aligned}
\end{equation}

For the architecture used in this study, where $n_0 = 9 + h$, $n_1 = 32$, $n_2 = 16$, $n_3 = 32$, and $n_4 = h$, this simplifies to the following.
\begin{equation}
    P = 1344 + 33h
\end{equation}
\subsubsection{Learning}

Weight optimization is performed using the Covariance Matrix Adaptation Evolution Strategy (CMA-ES). CMA-ES optimizes parameters by iteratively sampling candidate solutions from a multivariate normal distribution defined by a mean vector $\mu$, a global step size $\sigma$, and a covariance matrix $\mathbf{C}$. After evaluating sampled candidates, the mean is updated toward higher performing solutions, while the covariance matrix is adapted to capture correlations between parameters that lead to successful search directions. The step size controls the overall exploration scale and is adapted online to balance exploration and exploitation. Together, $(\mu, \sigma, \mathbf{C})$ define a learned search distribution that encodes both promising regions of the parameter space and parameter dependencies.

 We use the default parameters provided by the python-cmaes library with a population size of 20 and 1000 function evaluations per individual. Preliminary testing on handmade designs indicated that 1000 function evaluations were sufficient with minimal learning time.
\subsection{Lifetime Inheritance}

In CMA-ES, optimization proceeds by adapting a search distribution parameterized by a mean vector $\mu$, a global step size $\sigma$, and a covariance matrix $\mathbf{C}$. When an offspring is created, its morphology may differ from that of its parent. In this case, we transfer and transform the parent’s learned distribution parameters to account for the morphological change and to provide a warm start for optimization. This Lamarckian inheritance allows offspring to reuse information about promising regions of the parameter space and parameter correlations, in contrast to Darwinian evolution, where each offspring reinitializes $(\mu, \sigma, \mathbf{C})$.

\subsubsection{Covariance Matrix}

The covariance matrix $\mathbf{C}$ encodes correlations between controller parameters that have proven beneficial during the parent’s lifetime. When morphological changes alter the neural controller’s architecture (e.g., added or removed sensors or actuators), a correspondence mapping aligns overlapping parent and offspring parameters. The correspondence mapping defines a deterministic alignment between parent and offspring controller parameters based on shared morphological components, identifying which parameters represent the same functional elements across morphologies. The correlations between the shared parameters are preserved, as encoded by the CMA-ES covariance matrix $\mathbf{C}$, while the new parameters are initialized as uncorrelated dimensions:

\begin{equation}
C_o[i,j] =
\begin{cases}
C_p[i_p,j_p], & \text{if } w_i,w_j \text{ correspond to parent}\\
&\text{  weights} \\
\delta_{ij}, & \text{otherwise}
\end{cases}
\end{equation}

where $\delta_{ij}$ is the Kronecker delta, initializing new parameters with unit variance and zero covariance. This preserves learned dependencies in the search distribution while ensuring $\mathbf{C}_o$ remains symmetric and positive definite.

\subsubsection{Step Size and Mean Adaptation}

The mean vector $\mu$ represents the current center of the search distribution, while the step size $\sigma$ controls its overall scale. For inherited parameters, the offspring initializes its mean from the parent’s mean; newly introduced parameters are initialized at zero. To account for morphological change, the step size is blended with the initial step size based on the relative change in controller dimensionality:

\begin{align}
\sigma_o &= (1 - \alpha)\sigma_p + \alpha\sigma_{\text{init}}, \quad 
\alpha = \frac{|d_o - d_p|}{d_o}, \\
\mu_o[i] &=
\begin{cases}
\mu_p[i_p], & \text{if } w_i \text{ corresponds to a parent}\\
&\text{parameter} \\
0, & \text{otherwise.}
\end{cases}
\end{align}

\subsubsection{Parent Selection}

When there are multiple candidate parents, inheritance is taken from the parent with the most similar morphology, measured using tree edit distance. This maximizes the relevance of the inherited search distribution, particularly the covariance structure that captures parameter correlations.

\section{Experimental Setup}\label{sec:experimental_setup}

\subsection{Methodology and Procedure}
We employ a $2\times2$ factorial experimental design to systematically investigate the effects of inheritance mechanism and fitness criterion on evolutionary dynamics and outcomes. The inheritance mechanism factor comprises two methods: Darwinian inheritance, wherein offspring controllers undergo random reinitialization, and Lamarckian inheritance, wherein offspring inherit parental learned controllers. These mechanisms are operationalized as described in Section 3. The fitness criterion factor similarly comprises two levels: \textbf{pure locomotion fitness}, measured as translational distance, and \textbf{combined locomotion-novelty fitness}, incorporating both locomotion performance and morphological diversity.\\

Morphological novelty is quantified through vectors composed of morphological descriptors, following the methodology established in our prior work \cite{miras2018search}. Specifically, the morpholigical descriptors we use are: branching, limbs, length of limbs, coverage, joints, proportion, symmetry, size, all normalized between 0 and 1. For each individual $i$, novelty $N_i$ is computed as the Euclidean distance to its nearest neighbor in morphological descriptor space, thereby incentivizing exploration of diverse phenotypic configurations. The combined fitness function is formulated as:
\begin{equation}
f_{\text{combined}} = f_{\text{locomotion}}\cdot f_{\text{novelty}}
\end{equation}

Each experimental condition executes for 50 generations with a population size of 30 offspring per generation. Individual fitness evaluation requires 1,000 learning episodes, yielding $50 \text{ generations} \times 30 \text{ offspring} \times 1{,}000 \text{ evaluations} = 1{,}500{,}000$ simulation episodes, plus 30 initial population evaluations, totaling 1,530,000 time-steps per experimental run or 1,530 evaluations. With an estimated computational cost of approximately 1 minute per evaluation, each run requires approximately 48 compute-hours. We repeated each experiment with different seeds for 20 runs, generating data for 80 runs in total.

\subsection{Performance Metrics and Statistical Analysis}

Experimental outcomes are evaluated using a set of performance metrics. Population fitness progression is tracked via the mean and maximum fitness across generations to characterize evolutionary progress. 
To assess structural inheritance, parent–-offspring morphological similarity is quantified as the Euclidean distance between parental and offspring descriptors, indicating the degree to which morphology is preserved through reproduction. Initial locomotion performance is defined as the first fitness evaluation of the offspring, conducted immediately after inheriting the weight of the parents.
Comparisons of locomotion fitness Between-conditions of the final-generation (mean per run, 20 runs each) were performed using Welch’s t-test (two-sided, unequal variances assumed), while differences in initial fitness (Figures 3–4) were assessed using the Mann–Whitney U test (two-sided).
\section{Results and Discussion}\label{sec:results}

\begin{figure}
    \centering
    \includegraphics[width=0.9\linewidth]{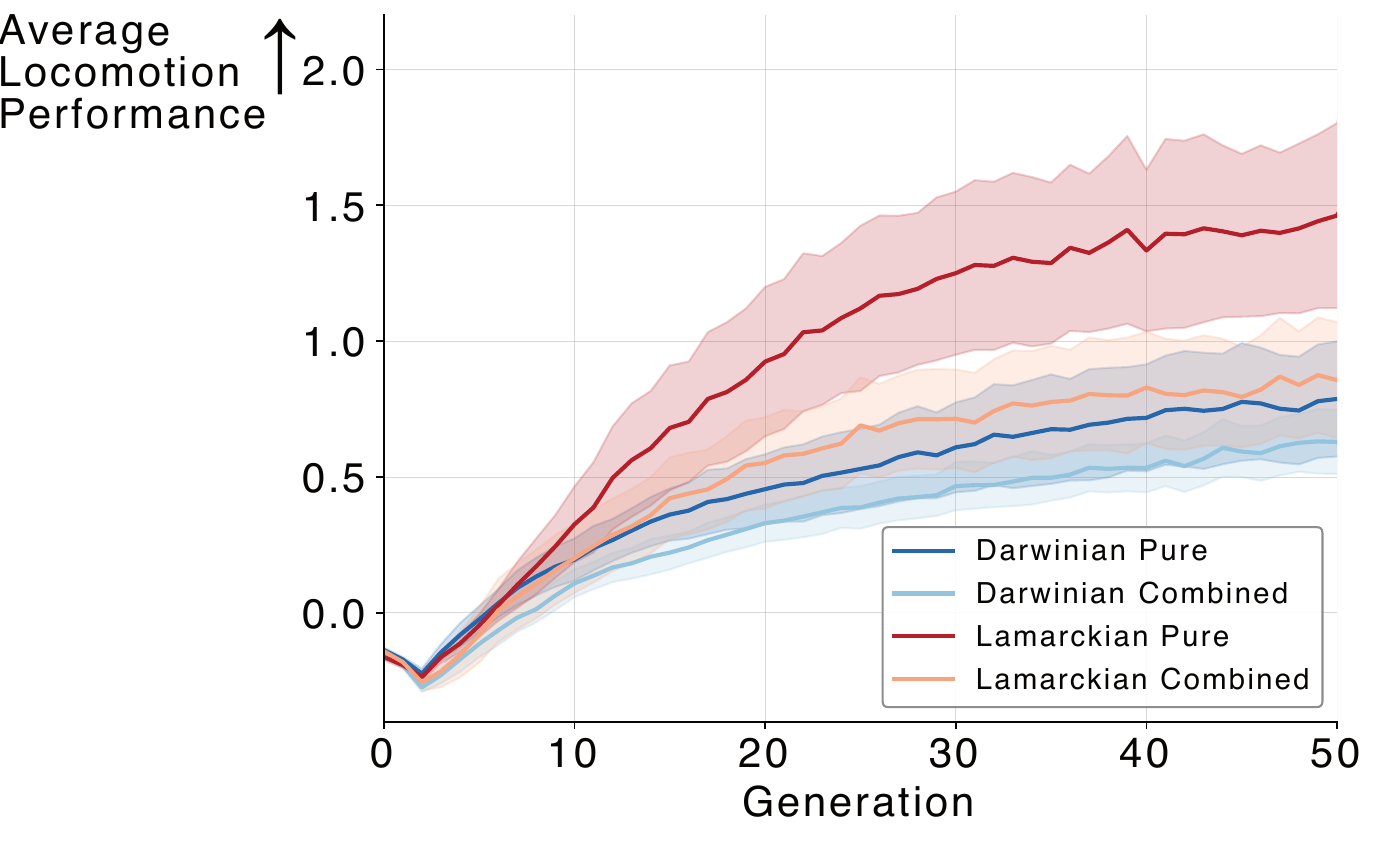}
    \caption{Average locomotion performance ($f_{locomotion}$) per generation. Higher is better.}
    \label{fig:locomotion_comparison}
\end{figure}

We start by analysing the performance of the four searches on the task of locomotion (Fig.~\ref{fig:locomotion_comparison}). All searches succeed in discovering robots that locomote effectively, with the rate of improvement decreasing substantially after approximately generation~30. The Lamarckian evolution outperforms Darwinian evolution in the search for locomotion performance (\textit{p=0.039}), consistent with prior work showing that Lamarckian inheritance accelerates learning. However, when morphological novelty is introduced as part of the fitness, the advantage of Lamarckian evolution in comparison to Darwinian evolution for locomotion performance is no longer significant (\textit{p=0.136}). While the Lamarckian system presents a significant reduction in locomotion performance under the pressure for morphological novelty (\textit{p=0.016}), this reduction in the Darwinian system under the same pressure is insignificant (\textit{p=0.071}).

\begin{figure}
    \centering
    \includegraphics[width=0.9\linewidth]{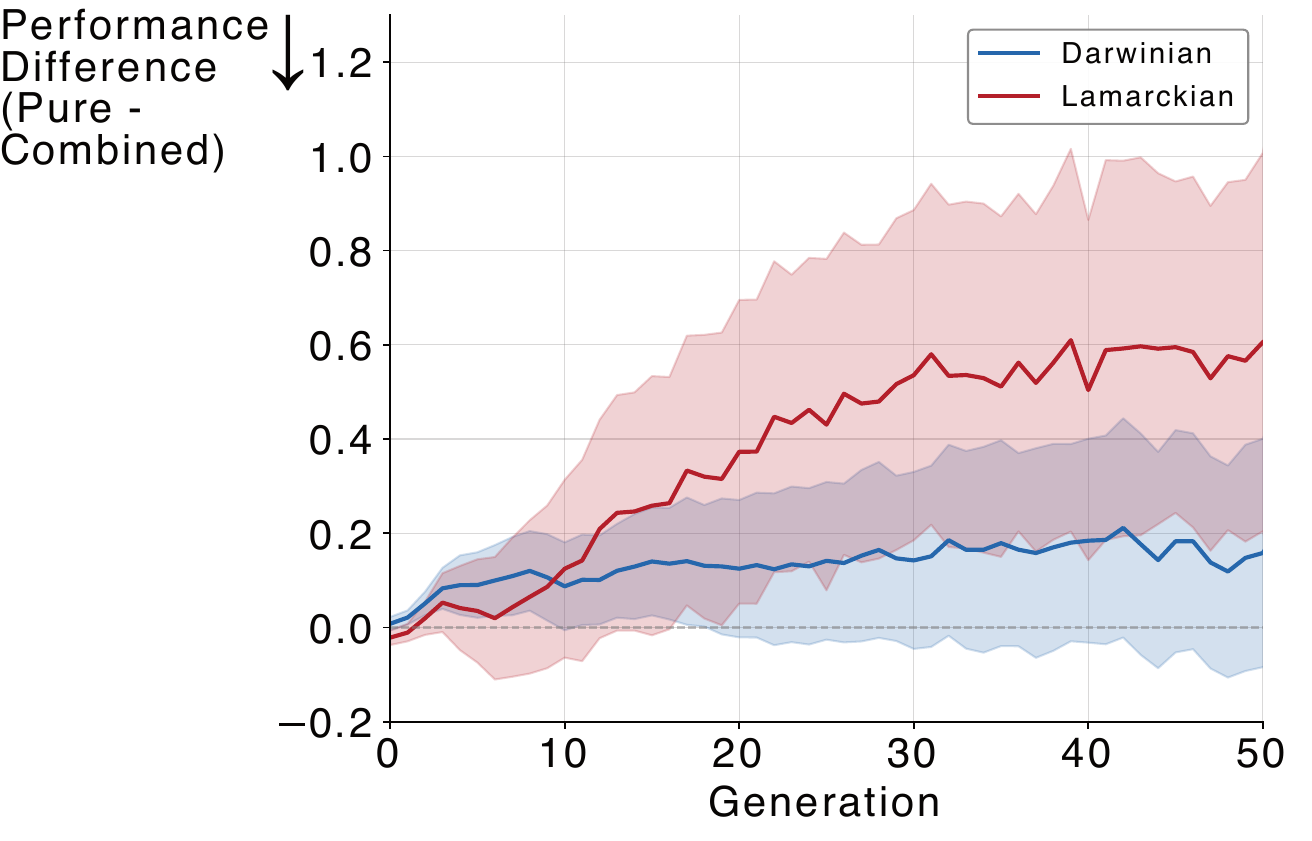}
    \caption{Average difference in locomotion performance. Lower is better. Performance differences between pure locomotion only and combined locomotion-novelty fitness, comparing Darwinian and Lamarckian setups. \textbf{The introduction of novelty has a more detrimental effect on Lamarckian evolution}.}
    \label{fig:locomotion_difference}
\end{figure}

Figure~\ref{fig:locomotion_difference} makes this effect explicit by showing the difference in performance between the searches that maximize locomotion and the searches that maximize for locomotion and novelty. The results directly support our hypothesis: \textit{the Lamarckian system suffers a substantially larger performance drop when novelty becomes part of the fitness, whereas the Darwinian system is almost unaffected}.


\begin{figure}
    \centering
    \includegraphics[width=0.9\linewidth]{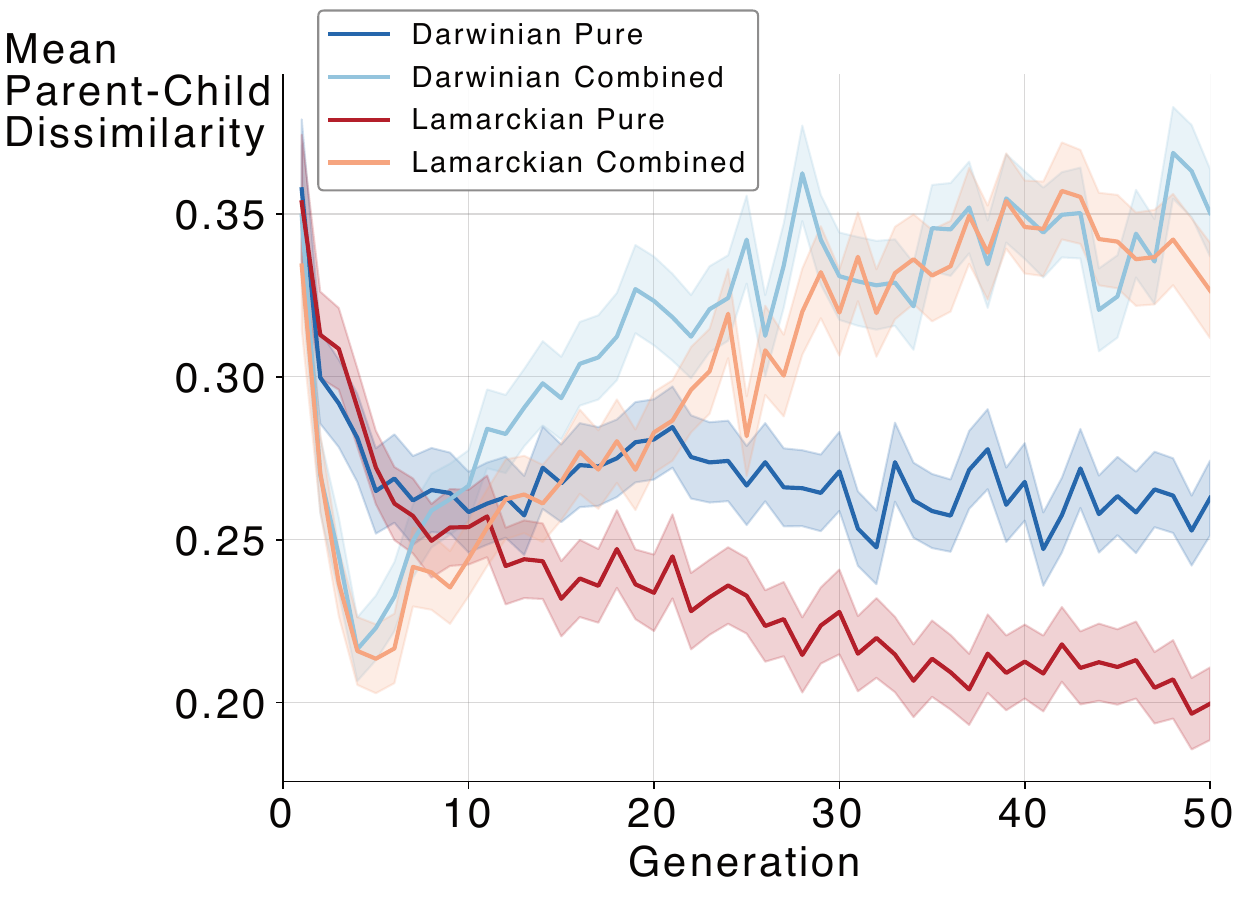}
    \caption{Mean parent–child morphological dissimilarity across generations (lower means more similar). Measured as the Euclidean distance between parental and offspring descriptors. \textbf{Lamarckian evolution exhibits a bias toward parent-offspring similarity, but this breaks down when novelty is introduced into the fitness function}.}
    \label{fig:mean_distance_over_generations}
\end{figure}

\begin{figure}
    \centering
    \includegraphics[width=\linewidth]{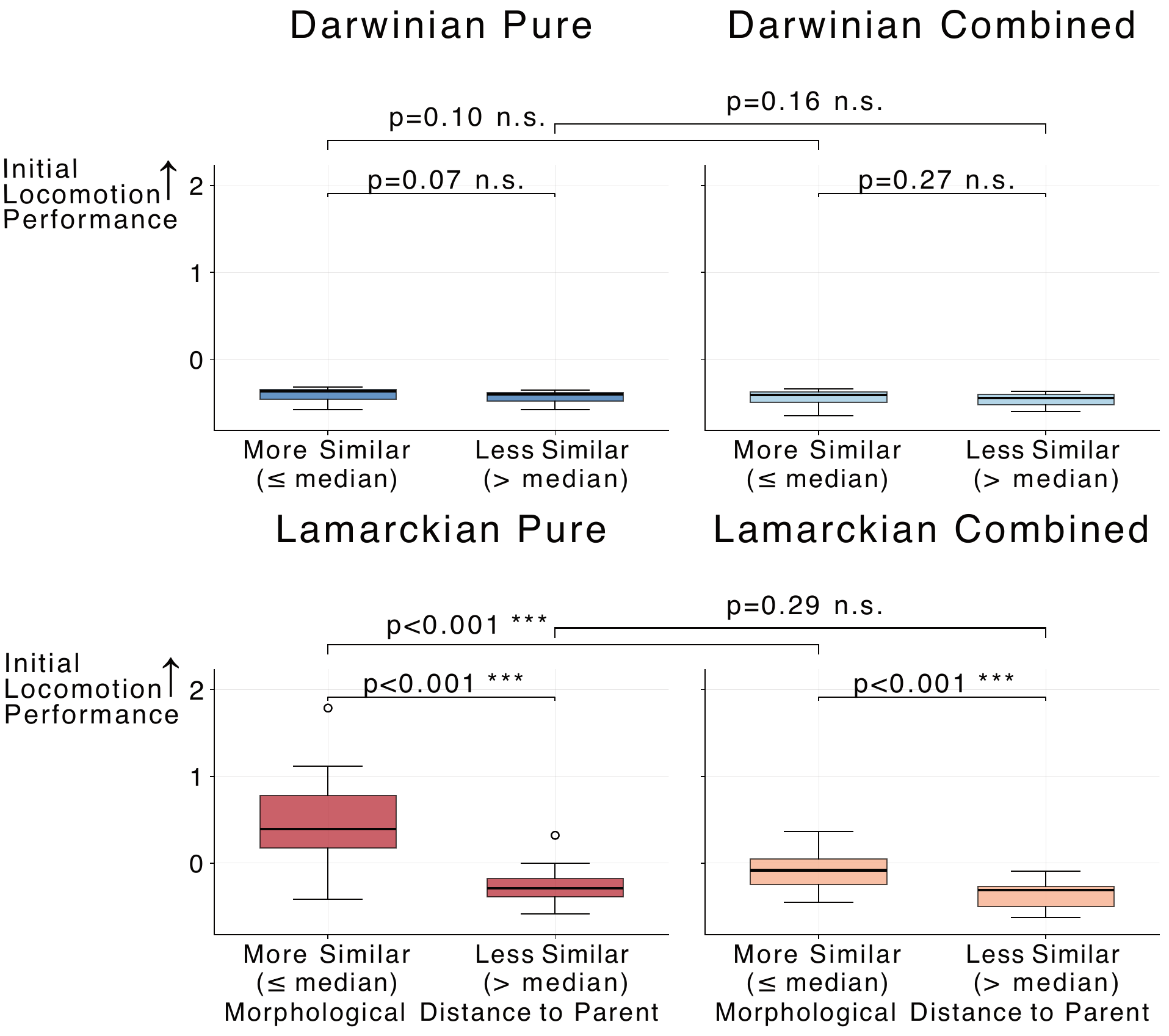}
    \caption{Initial locomotion performance within subgroups of parent-offspring morphological similarity. The two groups, more similar and less similar, are split using the median parent–offspring similarity of all individuals. \textbf{Under Lamarckian inheritance, greater parent-offspring similarity translates to higher initial locomotion performance}.}
    \label{fig:initfit_boxplot}
\end{figure}

Now that the hypothesis has been confirmed, we conduct a deeper analysis to evaluate whether the underlying intuition is valid: the Lamarckian system is more affected by novelty because it depends more heavily on similarity -- a dependence that follows from assuming no external pressure toward morphological diversity. If this intuition is correct, then three assumptions must hold, connected as follows:
 
\textit{(1)} The Lamarckian system `prefers' similar bodies more than the Darwinian system -- meaning that it works better with similar bodies (Fig. \ref{fig:initfit_boxplot}, comparisons between boxplots of the same color); therefore, it produces higher parent–offspring similarity when selecting only for performance (Fig.~\ref{fig:mean_distance_over_generations}, see red and dark blue curves). \textit{(2)} However, parent–offspring similarity is reduced by novelty (Fig.~\ref{fig:mean_distance_over_generations}, see orange and light blue curves). \textit{(3)} This reduction results in decreased locomotion performance of the offspring in the Lamarckian system immediately after controller inheritance, and this effect is more pronounced when parent–offspring similarity is higher (Fig.~\ref{fig:initfit_boxplot}, see comparisons among red and orange box-plots).

We examine the data supporting each of these assumptions in detail and find that all are upheld. First, we inspect the morphological similarity between parents and children and their evolution over time. Figure~\ref{fig:mean_distance_over_generations} shows the mean morphological similarity between parents and children for all conditions. In all experiments, similarity increases sharply during the initial generations. When novelty is incorporated, this initial increase is more pronounced, indicating stronger divergence of offspring from their parents; after this early phase, offspring become progressively more dissimilar. In the locomotion-only condition, however, offspring continue to become increasingly similar over time, but the Darwinian and Lamarckian systems differ in how convergence unfolds: the Lamarckian system produces greater parent-offspring similarity (assumption \textit{1}). This distinction disappears when novelty is introduced: similarity is reduced under both Darwinian and Lamarckian evolution (assumption 2), substantially narrowing the gap between the two approaches. 

Subsequently, we assess whether parent–offspring similarity modulates performance transference. Figure~\ref{fig:initfit_boxplot} shows the initial locomotion performance of the offspring immediately after transference -- this quantifies the benefit of the parent’s learned controller for the child’s initial performance. Under Darwinian evolution, the initial performance is largely insensitive to morphological similarity, while Lamarckian evolution shows a strong dependence. In the latter, offspring with greater parent-offspring similarity achieve substantially higher initial performance than less similar ones -- this reinforces the evidence that the Lamarckian system works better under greater parent-offspring similarity (assumption \textit{1}).  Additionally, the initial performance is significantly decreased in the novelty experiments, and this decrease is only significant for the sub-group with high parent-offspring similarity (assumption \textit{3}).

\begin{figure}
    \centering
    \includegraphics[width=\linewidth]{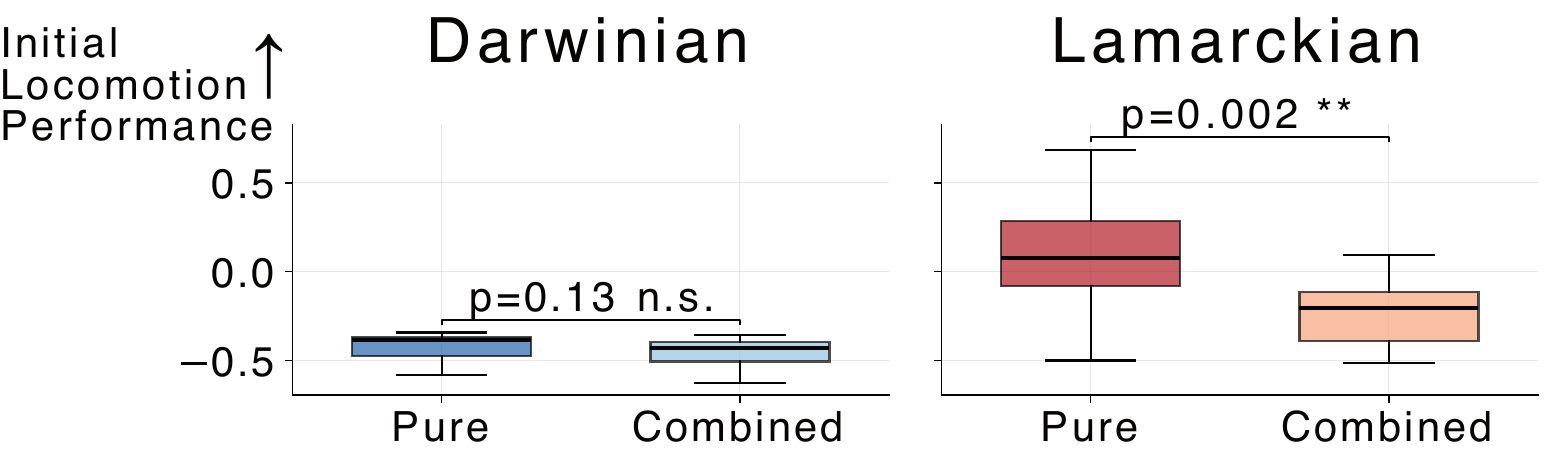}
    \caption{Comparison of initial locomotion performance. In this analysis, any individuals with performance above the maximum performance achieved in the combined experiments were removed from the sample; this aimed at preventing a bias that could be  introduced by the high performances values of the pure experiments. \textbf{Lamarckian evolution exhibits greater performance degradation when novelty is combined with a locomotion task, and this is due to decreased parent-offspring similarity}.} 
    \label{fig:initial_fitness_boxplot_pure_vs_combined}
\end{figure}

Since the pure locomotion fitness and the locomotion–novelty fitness yield distinct locomotion-performance distributions as a result of their respective selection pressures, one might question whether the performance differences between the novelty and pure settings are attributable solely to these distributions. If this were the case, the same type of difference would be observed consistently in both systems; however, it happens only in the Lamarckian system. To strengthen this evidence, we include an additional comparison in support of assumption \textit{3}. Since the pure setups achieve higher fitness than the combined setups, an effect that is more pronounced in the Lamarckian case, we cap individual fitness values in the pure experiments at the maximum fitness achieved in the corresponding combined experiments. Specifically, individuals in the Darwinian pure condition are retained only if their fitness does not exceed the maximum fitness observed in the Darwinian combined condition, and the same procedure is applied to the Lamarckian setups. This enables a direct comparison of initial fitness while avoiding bias introduced by differing performance ceilings. Figure \ref{fig:initial_fitness_boxplot_pure_vs_combined} shows that in the Darwinian system, there is no observable difference, whereas in the Lamarckian system, a significant difference in initial fitness emerges. 

In summary, these results show that the advantages of the Lamarckian system were hindered under the pressure towards morphological novelty, which in the present study was implemented as an explicit component of the fitness function. In non–open-ended evolutionary systems, where fitness is a user-defined function utilized for selection a priori rather than measured a posteriori as an outcome of natural selection, the fitness function serves as an abstraction of environmental pressures. Morphological novelty here, can be viewed as a controlled proxy for environmental forces that introduce transgenerational variability. In natural or dynamic environments, such forces may arise through frequency-dependent selection, niche differentiation, or environmental heterogeneity. What ultimately matters for controller inheritance is the effect on parent-offspring morphological continuity, regardless of whether variability originates from external environmental change or from diversity-promoting selection, as both disrupt the similarity assumption underlying Lamarckian inheritance.

\section{Conclusion}\label{sec:conclusion}

This paper investigated the benefits and limitations of Lamarckian evolution in comparison with Darwinian evolution in morphologically evolving robots that learn during their lifetime. We showed that while Lamarckian inheritance consistently improves performance when evolution favors morphological similarity, its advantage is reduced when novelty is explicitly rewarded. Darwinian evolution is largely unaffected. The key challenge arises when the pressure for morphological novelty reduces the similarity between parents and offspring, thereby affecting the conditions under which Lamarckian inheritance is effective.
These findings reveal an important limitation of Lamarckian evolution by exposing a fundamental trade-off between inheritance-based exploitation and diversity-driven exploration. Although Lamarckian evolution remains a powerful approach, its suitability depends on the conditions under which it is applied, and this work helps to delineate those conditions.

These observations open several avenues for future research aimed at identifying the regimes in which Lamarckian evolution is most effective. For example, it remains to be determined at what rate and magnitude of environmental change or variability introduction Lamarckian strategies cease to be advantageous. Future work could also explore adaptive inheritance mechanisms, such as similarity-aware transfer, partial controller reuse, or hybrid evolutionary schemes, to better harvest the benefits of Lamarckian learning with the demands of morphological diversity.

\section{Acknowledgments}
The authors would like to acknowledge that this work builds upon the ARIEL simulator, which is currently in the early stages of development. We sincerely thank the developers Jacopo Di Matteo, Áron Richárd Ferencz, Giannis Grigoriadis, and Lilly Schwarzenbach for their contributions to the development of this tool.

The research reported in this paper was supported by the European Commission Horizon project SPEAR (EU grant number 101119774, https://www.spear-robotics.com/).

\printbibliography

@article{jelisavcic2019lamarckian,
  title={Lamarckian Evolution of Simulated Modular Robots},
  author={Jelisavcic, Milan and Glette, Kyrre and Haasdijk, Evert and Eiben, A. E.},
  journal={Frontiers in Robotics and AI},
  volume={6},
  pages={9},
  year={2019},
  publisher={Frontiers},
  doi={10.3389/frobt.2019.00009}
}

@article{luo2025lamarckian,
  title={Lamarckian Inheritance Improves Robot Evolution in Dynamic Environments},
  author={Luo, Jie and Miras, Karine and Longhi, Carlo and Weissl, Oliver and Eiben, Agoston E.},
  journal={IEEE Transactions on Evolutionary Computation},
  year={2025},
  publisher={IEEE}
}

@inproceedings{miras2018search,
  author={Miras, Karine and Haasdijk, Evert and Glette, Kyrre and Eiben, A. E.},
  title={Search Space Analysis of Evolvable Robot Morphologies},
  booktitle={Applications of Evolutionary Computation: 21st International Conference, EvoApplications 2018, Parma, Italy, April 4-6, 2018, Proceedings},
  editor={Sim, Kevin and Kaufmann, Paul},
  series={Lecture Notes in Computer Science},
  volume={10784},
  pages={703--718},
  year={2018},
  publisher={Springer},
  address={Cham},
  doi={10.1007/978-3-319-77538-8_47}
}

@article{eiben2015grand,
  title={Grand Challenges for Evolutionary Robotics},
  author={Eiben, A. E. and Kernbach, Serge and Haasdijk, Evert},
  journal={Frontiers in Robotics and AI},
  volume={2},
  pages={4},
  year={2015},
  publisher={Frontiers}
}

@article{lehman2011abandoning,
  title={Abandoning Objectives: Evolution Through the Search for Novelty Alone},
  author={Lehman, Joel and Stanley, Kenneth O.},
  journal={Evolutionary Computation},
  volume={19},
  number={2},
  pages={189--223},
  year={2011},
  publisher={MIT Press}
}

@article{pugh2016quality,
  title={Quality Diversity: A New Frontier for Evolutionary Computation},
  author={Pugh, Justin K. and Soros, Lisa B. and Stanley, Kenneth O.},
  journal={Frontiers in Robotics and AI},
  volume={3},
  pages={40},
  year={2016},
  publisher={Frontiers}
}

@article{cully2015robots,
  title={Robots that Can Adapt Like Animals},
  author={Cully, Antoine and Clune, Jeff and Tarapore, Danesh and Mouret, Jean-Baptiste},
  journal={Nature},
  volume={521},
  number={7553},
  pages={503--507},
  year={2015},
  publisher={Nature Publishing Group}
}

@article{auerbach2014environmental,
  title={Environmental Influence on the Evolution of Morphological Complexity in Machines},
  author={Auerbach, Joshua E. and Bongard, Josh C.},
  journal={PLOS Computational Biology},
  volume={10},
  number={1},
  pages={e1003399},
  year={2014},
  publisher={Public Library of Science}
}

@article{cheney2013unshackling,
  title={Unshackling Evolution: Evolving Soft Robots with Multiple Materials and a Powerful Generative Encoding},
  author={Cheney, Nick and MacCurdy, Robert and Clune, Jeff and Lipson, Hod},
  journal={ACM SIGEVOlution},
  volume={7},
  number={1},
  pages={11--23},
  year={2013},
  publisher={ACM New York, NY, USA}
}

@inproceedings{jelisavcic2017benefits,
  title={Benefits of Lamarckian Evolution for Morphologically Evolving Robots},
  author={Jelisavcic, Milan and Kiesel, Rafael and Glette, Kyrre and Haasdijk, Evert and Eiben, A. E.},
  booktitle={Proceedings of the Genetic and Evolutionary Computation Conference Companion},
  pages={65--66},
  year={2017},
  organization={ACM}
}

@inproceedings{jelisavcic2017analysis,
  title={Analysis of Lamarckian Evolution in Morphologically Evolving Robots},
  author={Jelisavcic, Milan and Kiesel, Rafael and Glette, Kyrre and Haasdijk, Evert and Eiben, A. E.},
  booktitle={Artificial Life Conference Proceedings},
  pages={214--221},
  year={2017},
  organization={MIT Press}
}

@inproceedings{jelisavcic2018morphological,
  title={Morphological Attractors in Darwinian and Lamarckian Evolutionary Robot Systems},
  author={Jelisavcic, Milan and Miras, Karine and Eiben, A. E.},
  booktitle={2018 IEEE Symposium Series on Computational Intelligence (SSCI)},
  pages={859--866},
  year={2018},
  organization={IEEE}
}

@inproceedings{luo2023comparative,
  title={A Comparative Study of Brain Reproduction Methods for Morphologically Evolving Robots},
  author={Luo, Jie and Longhi, Carlo and Eiben, Agoston E.},
  booktitle={Artificial Life Conference Proceedings},
  volume={2023},
  number={1},
  pages={44},
  year={2023},
  organization={MIT Press}
}

@inproceedings{luo2023comparison,
  title={A Comparison of Controller Architectures and Learning Mechanisms for Arbitrary Robot Morphologies},
  author={Luo, Jie and Tomczak, Jakub M. and Miras, Karine and Eiben, Agoston E.},
  booktitle={2023 IEEE Symposium Series on Computational Intelligence (SSCI)},
  pages={1518--1525},
  year={2023},
  organization={IEEE}
}

@article{luo2023enhancing,
  title={Enhancing Robot Evolution Through Lamarckian Principles},
  author={Luo, Jie and Miras, Karine and Tomczak, Jakub M. and Eiben, Agoston E.},
  journal={Scientific Reports},
  volume={13},
  number={1},
  pages={21109},
  year={2023},
  publisher={Nature Publishing Group}
}

@article{mingo2013lamarckism,
  title={Investigations into Lamarckism, Baldwinism and Local Search in Grammatical Evolution Guided by Reinforcement},
  author={Mingo, Jack Mario and Aler, Ricardo and Maravall, Dar{\'\i}o and de Lope, Javier},
  journal={Computing and Informatics},
  volume={32},
  number={3},
  pages={595--627},
  year={2013}
}

@article{sasaki2000comparison,
  title={Comparison Between Lamarckian and Darwinian Evolution on a Model Using Neural Networks and Genetic Algorithms},
  author={Sasaki, Takahiro and Tokoro, Mario},
  journal={Knowledge and Information Systems},
  volume={2},
  number={2},
  pages={201--222},
  year={2000},
  publisher={Springer}
}

@inproceedings{sharifi2024lcodeepneat,
  title={Developing Convolutional Neural Networks Using a Novel Lamarckian Co-Evolutionary Algorithm},
  author={Sharifi, Zaniar and Soltanian, Khabat and Amiri, Ali},
  booktitle={13th International Conference on Computer and Knowledge Engineering (ICCKE)},
  year={2023},
  month={11},
  organization={Ferdowsi University of Mashhad},
  address={Mashhad, Iran}
}

@inproceedings{tack2024lamarckian,
  title={Lamarckian Co-design of Soft Robots via Transfer Learning},
  author={Tack, Nathan A. M. and Yamazaki, Yudai and Iida, Fumiya},
  booktitle={Proceedings of the Genetic and Evolutionary Computation Conference},
  pages={129--137},
  year={2024},
  organization={ACM}
}

\end{document}